\documentclass[10pt, a4paper]{article}
\usepackage{ltc05}
\usepackage[T1]{fontenc}
\usepackage[utf8]{inputenc}

\usepackage{amssymb,amsmath}
\DeclareMathOperator{\ReLU}{ReLU}
\DeclareMathOperator{\D}{D}
\DeclareMathOperator{\CD}{CD}
\DeclareMathOperator{\LD}{LD}
\DeclareMathOperator{\V}{V}
\DeclareMathOperator{\perplexity}{perplexity}

\title{Evaluation of basic modules for isolated spelling error correction in Polish texts}

\name{Szymon Rutkowski$^{\ast}$} 

\address{ $^{\ast}$University of Warsaw \\
               Krakowskie Przedmieście 26/28, 00-927 Warsaw, Poland \\
               szymon@szymonrutkowski.pl}

\abstract{Spelling error correction is an important problem in natural language processing, as a prerequisite for good performance in downstream tasks as well as an important feature in user-facing applications. For texts in Polish language, there exist works on specific error correction solutions, often developed for dealing with specialized corpora, but not evaluations of many different approaches on big resources of errors. We begin to address this problem by testing some basic and promising methods on PlEWi, a corpus of annotated spelling extracted from Polish Wikipedia. These modules may be further combined with appropriate solutions for error detection and context awareness. Following our results, combining edit distance with cosine distance of semantic vectors may be suggested for interpretable systems, while an LSTM, particularly enhanced by ELMo embeddings, seems to offer the best raw performance.}

\begin{document}

\maketitleabstract

\section{Introduction}

Spelling error correction is a fundamental NLP task. Most language processing applications benefit greatly from being provided clean texts for their best performance. Human users of computers also often expect competent help in making spelling of their texts correct.

Because of the lack of tests of many common spelling correction methods for Polish, it is useful to establish how they perform in a simple scenario. We constrain ourselves to the pure task of isolated correction of non-word errors. They are traditionally separated in error correction literature \cite{kukich1992techniques}. Non-word errors are here incorrect word forms that not only differ from what was intended, but also do not constitute another, existing word themselves. Much of the initial research on error correction focused on this simple task, tackled without means of taking the context of the nearest words into account.

It is true that, especially in the case of neural networks, it is often possible and desirable to combine problems of error detection, correction and context awareness into one task trained with a supervised training procedure. In language correction research for English language also grammatical and regular spelling errors have been treated uniformly with much success \cite{Ge2018ReachingHP}.

However, when more traditional methods are used, because of their predictability and interpretability for example, one can mix and match various approaches to dealing with the subproblems of detection, correction and context handling (often equivalent to employing some kind of a language model). We call it a modular approach to building spelling error correction systems. There is recent research where this paradigm was applied, interestingly, to convolutional networks trained separately for various subtasks \cite{dronen}. In similar setups it is more useful to assess abilities of various solutions in isolation. The exact architecture of a spelling correction system should depend on characteristics of texts it will work on.

Similar considerations eliminated from our focus handcrafted solutions for the whole spelling correction pipeline, primarily the LanguageTool \cite{milkowski}. Its performance in fixing spelling of Polish tweets was already tested \cite{ogr:kop:17}. For our purposes it would be given an unfair advantage, since it is a rule-based system making heavy use of words in context of the error.

\section{Problems of spelling correction for Polish}

Published work on language correction for Polish dates back at least to 1970s, when simplest Levenshtein distance solutions were used for cleaning mainframe inputs \cite{subieta1976,subieta1985simple}. Spelling correction tests described in literature have tended to focus on one approach applied to a specific corpus. Limited examples include works on spellchecking mammography reports and tweets \cite{mammografia,ogr:kop:17}. These works emphasized the importance of tailoring correction systems to specific problems of corpora they are applied to. For example, mammography reports suffer from poor typing, which in this case is a repetitive work done in relative hurry. Tweets, on the other hand, tend to contain emoticons and neologisms that can trick solutions based on rules and dictionaries, such as LanguageTool. The latter is, by itself, fairly well suited for Polish texts, since a number of extensions to the structure of this application was inspired by problems with morphology of Polish language \cite{milkowski}.

These existing works pointed out more general, potentially useful qualities specific to spelling errors in Polish language texts. It is, primarily, the problem of leaving out diacritical signs, or, more rarely, adding them in wrong places. This phenomenon stems from using a variant of the US keyboard layout, where combinations of \texttt{AltGr} with some alphabetic keys produces characters unique to Polish. When the user forgets or neglects to press the AltGr key, typos such as writing \textit{*olowek} instead of \textit{ołówek} appear. In fact, \cite{ogr:kop:17} managed to get substantial performance on Twitter corpus by using this ''diacritical swapping'' alone.

\section{Methods}

\subsection{Baseline methods}

The methods that we evaluated are baselines are the ones we consider to be basic and with moderate potential of yielding particularly good results. Probably the most straightforward approach to error correction is selecting known words from a dictionary that are within the smallest edit distance from the error. We used the Levenshtein distance metric \cite{levenshtein1966bcc} implemented in Apache Lucene library \cite{pylucene}. It is a version of edit distance that treats deletions, insertions and replacements as adding one unit distance, without giving a special treatment to character swaps. The SGJP -- Grammatical Dictionary of Polish \cite{sgjp} was used as the reference vocabulary.

Another simple approach is the aforementioned diacritical swapping, which is a term that we introduce here for referring to a solution inspired by the work of \cite{ogr:kop:17}. Namely, from the incorrect form we try to produce all strings obtainable by either adding or removing diacritical marks from characters. We then exclude options that are not present in SGJP, and select as the correction the one within the smallest edit distance from the error. It is possible for the number of such diacritically-swapped options to become very big. For example, the token \textit{Modlin-Zegrze-Pultusk-Różan-Ostrołęka-Łomża-Osowiec} (taken from PlEWi corpus of spelling errors, see below) can yield over \(2^{29}=536,870,912\) states with this method, such as \textit{Módłiń-Żęgrzę-Pułtuśk-Roźąń-Óśtróleką-Lómzą-Óśówięć}. The actual correction here is just fixing the \textit{ł} in \textit{Pułtusk}. Hence we only try to correct in this way tokens that are shorter than 17 characters.

\subsection{Vector distance}

A promising method, adapted from work on correcting texts by English language learners \cite{NAGATA2017474}, expands on the concept of selecting a correction nearest to the spelling error according to some notion of \textit{distance}. Here, the Levenshtein distance is used in a weighted sum to cosine distance between word vectors. This is based on the observation that trained vectors models of distributional semantics contain also representations of spelling errors, if they were not pruned. Their representations tend to be similar to those of their correct counterparts. For example, the token \textit{enginir} will appear in similar contexts as \textit{engineer}, and therefore will be assigned a similar vector embedding.

The distance between two tokens $a$ and $b$ is thus defined as

\[
   \D(a, b) = \frac{\LD(a,b) + \CD(\V(a), \V(b))}{2}.
   \]

   Here $\LD$ is just Levenshtein distance between strings, and $\CD$ -- cosine distance between vectors. $\V(a)$ denotes the word vector for $a$. Both distance metrics are in our case roughly in the range [0,1] thanks to the scaling of edit distance performed automatically by Apache Lucene. We used a pretrained set of word embeddings of Polish \cite{mykowiecka_wektory}, obtained with the flavor word2vec procedure using skipgrams and negative sampling \cite{mikolov}.

\subsection{Recurrent neural networks}

Another powerful approach, if conceptually simple in linguistic terms, is using a character-based recurrent neural network. Here, we test uni- and bidirectional Long Short-Term Memory networks \cite{Hochreiter:1997:LSM:1246443.1246450} that are fed characters of the error as their input and are expected to output its correct form, character after character. This is similar to traditional solutions conceptualizing the spelling error as a chain of characters, which are used as evidence to predict the most likely chain of replacements (original characters). This was done with n-gram methods, Markov chains and other probabilistic models \cite{araki1994evaluation}. Since nowadays neural networks enjoy a large awareness as an element of software infrastructure, with actively maintained packages readily available, their evaluation seems to be the most practically useful. We used the PyTorch \cite{Paszke2017AutomaticDI} implementation of LSTM in particular.

The bidirectional version \cite{Schuster:1997:BRN:2198065.2205129} of LSTM reads the character chains forward and backwards at the same time. Predictions from networks running in both directions are averaged.

In order to provide the network an additional, broad picture peek at the whole error form we also evaluated a setup where the internal state of LSTM cells, instead of being initialized randomly, is computed from an ELMo embedding \cite{DBLP:journals/corr/abs-1802-05365} of the token. The ELMo embedder is capable of integrating linguistic information carried by the whole form (probably often not much in case of errors), as well as the string as a character chain. The latter is processed with a convolutional neural network. How this representation is constructed is informed by the whole corpus on which the embedder was trained. The pretrained ELMo model that we used \cite{Che2018TowardsBU} was trained on Wikipedia and Common Crawl corpora of Polish.

The ELMo embedding network outputs three layers as matrices, which are supposed to reflect subsequent compositional layers of language, from phonetic phenomena at the bottom to lexical ones at the top. A weighted sum of these layers is computed, with weights trained along with the LSTM error-correcting network. Then we apply a trained linear transformation, followed by $\ReLU$ non-linearity:
\[
   \ReLU(x) = \max (0, x)
\]
(applied cellwise) in order to obtain the initial setting of parameters for the main LSTM. Our ELMo-augmented LSTM is bidirectional.

\begin{table*}[ht]
\begin{center}
\begin{tabular}{ccccc}
\hline
\textbf{Method}         &       \textbf{Accuracy} & \textbf{Perplexity} & \textbf{Loss (train)} & \textbf{Loss (test)} \\
\hline
Edit distance       &      0.3453 & - & - & - \\
Diacritic swapping     &      0.2279 & - & - & - \\
Vector distance      &      0.3945 & - & - & - \\
LSTM-1 net              &      0.4183 & \textbf{907} & 0.3 & 0.41 \\
LSTM-2 net              &      0.6634 & 11182 & 0.1 & \textbf{0.37} \\
LSTM-ELMo net           &      \textbf{0.6818} & 706166 & \textbf{0.07} & 0.38 \\
\hline
\end{tabular}
\caption{Test results for all the methods used. The loss measure is cross-entropy.}\label{table:results}
\end{center}
\end{table*}

\begin{table}
\begin{center}
\begin{tabular}{ccc}
\hline
\textbf{Layer I} & \textbf{Layer II} & \textbf{Layer III} \\
\hline
0.036849 & 0.08134 & 0.039395 \\
\hline
\end{tabular}
\caption{Discovered optimal weights for summing layers of ELMo embedding for initializing an error-correcting LSTM. The layers are numbered from the one that directly processes character and word input to the most abstract one.}\label{table:elmo_weights}
\end{center}
\end{table}

\section{Experimental setup}

PlEWi \cite{grundkiewicz:automatic} is an early version of WikEd \cite{Grundkiewicz2014TheWE} error corpus, containing error type annotations allowing us to select only non-word errors for evaluation. Specifically, PlEWi supplied 550,755 [error, correction] pairs, from which 298,715 were unique. The corpus contains data extracted from histories of page versions of Polish Wikipedia. An algorithm designed by the corpus author determined where the changes were correcting spelling errors, as opposed to expanding content and disagreements among Wikipedia editors.

The corpus features texts that are descriptive rather than conversational, contain relatively many proper names and are more likely to have been at least skimmed by the authors before submitting for online publication. Error cases provided by PlEWi are, therefore, not a balanced representation of spelling errors in written Polish language. PlEWi does have the advantage of scale in comparison to existing literature, such as \cite{ogr:kop:17} operating on a set of only 740 annotated errors in tweets.

All methods were tested on a test subset of 25\% of cases, with 75\% left for training (where needed) and 5\% for development.

The methods that required training -- namely recurrent neural networks -- had their loss measured as cross-entropy loss measure between correct character labels and predictions. This value was minimized with Adam algorithm \cite{DBLP:journals/corr/KingmaB14}. The networks were trained for 35 epochs.

\section{Results}

The experimental results are presented in Table \ref{table:results}. Diacritic swapping showed a remarkably poor performance, despite promising mentions in existing literature. This might be explained by the already mentioned feature of Wikipedia edits, which can be expected to be to some degree self-reviewed before submission. This can very well limit the number of most trivial mistakes.

On the other hand, the vector distance method was able to bring a discernible improvement over pure Levenshtein distance, comparable even with the most basic LSTM. It is possible that assigning more fine-tuned weights to edit distance and semantic distance would make the quality of predictions even higher. The idea of using vector space measurements explicitly can be also expanded if we were to consider the problem of contextualizing corrections. For example, the semantic distance of proposed corrections to the nearest words is likely to carry much information about their appropriateness. Looking from another angle, searching for words that seem semantically off in context may be a good heuristic for detecting errors that are not nonword (that is, they lead to wrong forms appearing in text which are nevertheless in-vocabulary).

The good performance of recurrent network methods is hardly a surprise, given observed effectiveness of neural networks in many NLP tasks in the recent decade. It seems that bidirectional LSTM augmented with ELMo may already hit the limit for correcting Polish spelling errors without contextual information. While it improves accuracy in comparison to LSTM initialized withrandom noise, it makes the test cross-entropy slightly worse, which hints at overfitting. The perplexity measures actually increase sharply for more sophisticated architectures. Perplexity should show how little probability is assigned by the model to true answers. We measure it as

\[
\perplexity(P, x) = 2^{-\frac{1}{N}\sum_{i \leqslant N} \log P(x_i)},
\]

where $x$ is a sequence of $N$ characters, forming the correct version of the word, and $P(x_i)$ is the estimated probability of the $i$th character, given previous predicted characters and the incorrect form. The observed increase of perplexity for increasingly accurate models is most likely due to more refined predicted probability distributions, which go beyond just assigning the bulk of probability to the best answer.

Interesting insights can be gained from weights assigned by optimization to layers of ELMo network, which are taken as the word form embedding (Table \ref{table:elmo_weights}). The first layer, and the one that is nearest to input of the network, is given relatively the least importance, while the middle one dominates both others taken together. This suggests that in error correction, at least for Polish, the middle level of morphemes and other characteristic character chunks is more important than phenomena that are low-level or tied to some specific words. This observation should be taken into account in further research on practical solutions for spelling correction.

\section{Conclusion}

Among the methods tested the bidirectional LSTM, especially initialized by ELMo embeddings, offers the best accuracy and raw performance. Adding ELMo to a straightforward PyTorch implementation of LSTM may be easier now than at the time of performing our tests, as since then the authors of ELMoForManyLangs package \cite{Che2018TowardsBU} improved their programmatic interface. However, if a more interpretable and explainable output is required, some version of vector distance combined with edit distance may be the best direction. It should be noted that this method produces multiple candidate corrections with their similarity scores, as opposed to only one ``best guess`` correction that can be obtained from a character-based LSTM. This is important in applications where it is up to humans to the make the final decision, and they are only to be aided by a machine.

It is desirable for further reasearch to expand the corpus material into a wider and more representative set of texts. Nevertheless, the solution for any practical case has to be tailored to its characteristic error patterns. Works on language correction for English show that available corpora can be ''boosted'' \cite{Ge2018ReachingHP}, i.e. expanded by generating new errors consistent with a generative model inferred from the data. This may greatly aid in developing models that are dependent on learning from error corpora.

A deliberate omission in this paper are the elements accompanying most real-word error correction solutions. Some fairly obvious approaches to integrating evidence from context include n-grams and Markov chains, although the possibility of using measurements in spaces of semantic vectors was already mentioned in this article. Similarly, non-word errors can be easily detected with comparing tokens against reference vocabulary, but in practice one should have ways of detecting mistakes masquerading as real words and fixing bad segmentation (tokens that are glued together or improperly separated). Testing how performant are various methods for dealing with these problems in Polish language is left for future research.

\bibliographystyle{ltc05}
\bibliography{spellcor} 

\end{document}